\documentclass[letterpaper, 10 pt, journal, twoside]{IEEEtran}

\usepackage{times}
\usepackage{amsmath,amsfonts}
\usepackage{algorithmic}
\usepackage{algorithm}
\usepackage{array}
\usepackage[font=small]{caption}
\usepackage{subcaption}
\usepackage{textcomp}
\usepackage{stfloats}
\usepackage{url}
\usepackage{verbatim}
\usepackage{graphicx}
\usepackage{cite}
\usepackage{multicol}
\usepackage[hidelinks]{hyperref}
\usepackage{glossaries}
\usepackage{xcolor}
\usepackage{soul}

\usepackage{booktabs}
\usepackage[table]{xcolor}
\definecolor{RowLight}{gray}{1.0}
\definecolor{RowDark}{gray}{0.85}

\newcommand{\revf}[1]{\textcolor{black}{#1}}
\newcommand{\revs}[1]{\textcolor{black}{#1}}
\newacronym{mpc}{MPC}{Model Predictive Control}
\newacronym{rti}{RTI}{Real-Time Iteration}
\newacronym{ocp}{OCP}{Optimal Control Problem}
\newacronym{srbd}{SRBD}{Single Rigid Body Dynamics}
\newacronym{qp}{QP}{Quadratic Programming}
\newacronym{admm}{ADMM}{Alternating Direction Method of Multipliers}
\newacronym{nlp}{NLP}{Nonlinear Programming}
\newacronym{wb}{WB}{Whole Body}
\newacronym{grf}{GRF}{Ground Reaction Forces}
\newacronym{sqp}{SQP}{Sequential quadratic Programming}
\newacronym{ddp}{DDP}{Differential Dynamic Programming}
\newacronym{ilqr}{iLQR}{Iterative Linear Quadratic Regulator}
\newacronym{mhe}{MHE}{Moving Horizon Estimator}
\newacronym{rl}{RL}{Reinforcement Learning}
\newacronym{cvf}{CVF}{Conditional Value Function}

\usepackage{xcolor}
\definecolor{lgray}{gray}{0.30}
\usepackage{eso-pic}
\usepackage{hyperref}
\AddToShipoutPictureBG*{%
  \AtPageUpperLeft{%
    \setlength{\unitlength}{1mm}%
    \put(-18.5,-9){\makebox(\paperwidth,0)[c]{\parbox{0.8\textwidth}{IEEE Robotics and Automation Letters, 2026, vol. 11, no. 1, pp. 1010-1017}}}
  }
}
\AddToShipoutPictureBG*{%
  \AtPageUpperLeft{%
    \setlength{\unitlength}{1mm}%
    \put(-9.5,-14){\makebox(\paperwidth,0)[c]{\parbox{0.9\textwidth}{DOI: \href{https://ieeexplore.ieee.org/document/11248841}{10.1109/LRA.2025.3632610}}}}
  }
}

\begin{document}

\title{Primal-Dual iLQR for GPU-Accelerated Learning and Control \\ in Legged Robots}
\author{Lorenzo Amatucci$^{*,1,2}$, João Sousa-Pinto$^{*,3}$, Giulio Turrisi$^1$, Dominique Orban$^4$,  \\ Victor Barasuol$^1$, Claudio Semini$^1$
\thanks{\textsuperscript{*} Equal Contribution}
\thanks{\textsuperscript{1} Dynamic Legged Systems Laboratory, Istituto Italiano di Tecnologia (IIT), Genova, Italy. E-mail: name.lastname@iit.it} \thanks{\textsuperscript{2} Dipartimento di Informatica, Bioingegneria, Robotica e Ingegneria dei Sistemi (DIBRIS), Università di Genova, Genova, Italy.}
\thanks{\textsuperscript{3} This work was done while João Sousa-Pinto was at Apple.}
\thanks{\textsuperscript{4} GERAD and Department of Mathematics and Industrial Engineering, Polytechnique Montréal,
Montreal, Canada}
}


\maketitle
\begin{abstract}
This paper introduces a novel Model Predictive Control (MPC) implementation for legged robot locomotion that leverages GPU parallelization. Our approach enables both temporal and state-space parallelization by incorporating a parallel associative scan to solve the primal-dual Karush-Kuhn-Tucker (KKT) system. In this way, the optimal control problem is solved in $\mathcal{O}(\log^2(n)\log{N} + \log^2(m))$ complexity, instead of $\mathcal{O}(N(n + m)^3)$, where $n$, $m$, and $N$ are the dimension of the system state, control vector, and the length of the prediction horizon.
We demonstrate the advantages of this implementation over two state-of-the-art solvers (acados and crocoddyl), achieving up to a 60\% improvement in runtime for Whole Body Dynamics (WB)-MPC and a 700\% improvement for Single Rigid Body Dynamics (SRBD)-MPC when varying the prediction horizon length. The presented formulation scales efficiently with the problem state dimensions as well, enabling the definition of a centralized controller for up to 16 legged robots that can be computed in less than 25 ms.
Furthermore, thanks to the JAX implementation, the solver supports large-scale parallelization across multiple environments, allowing the possibility of performing learning with the MPC in the loop directly in GPU. The code associated with this work can be found at \url{https://github.com/iit-DLSLab/mpx}.
\end{abstract}

\IEEEpeerreviewmaketitle



\section{Introduction}  
\IEEEPARstart{A}{mong} the many well-known control approaches available, \gls{mpc} has proven to be highly effective in generating and controlling complex dynamic behaviors in robotic systems, especially legged robots, as shown by \cite{Mpc} and \cite{Perceptive_based_MPC}. At the core of an \gls{mpc} is the transcription of a task we want the robot to perform into an \gls{ocp} and the ability to solve it fast enough to be used in a closed-loop controller. 

One of the most widely used methods in robotics for solving such \gls{ocp} is \gls{ddp}, which gained renewed attention through the work of Todorov et al.\cite{ilqr}, who introduced a variant known as \gls{ilqr}. \gls{ilqr} discards the second-order terms of the dynamics in the Hessian, sacrificing the local quadratic convergence of \gls{ddp} in favor of faster update rates. However, a key limitation of this method is its single-shooting nature, which requires a feasible initial guess as a starting point and often exhibits critical numerical issues. \cite{GNSM} overcame such limitation presenting a multiple shooting variant of the \gls{ilqr} algorithm.
More recently, \cite{crocoddyl} proposed a feasibility-driven multiple shooting approach to \gls{ddp} with notable results also on real hardware \cite{talosWholeBody}. 
\begin{figure}[!t]
     \centering
         \centering
         \includegraphics[width=0.45\textwidth]{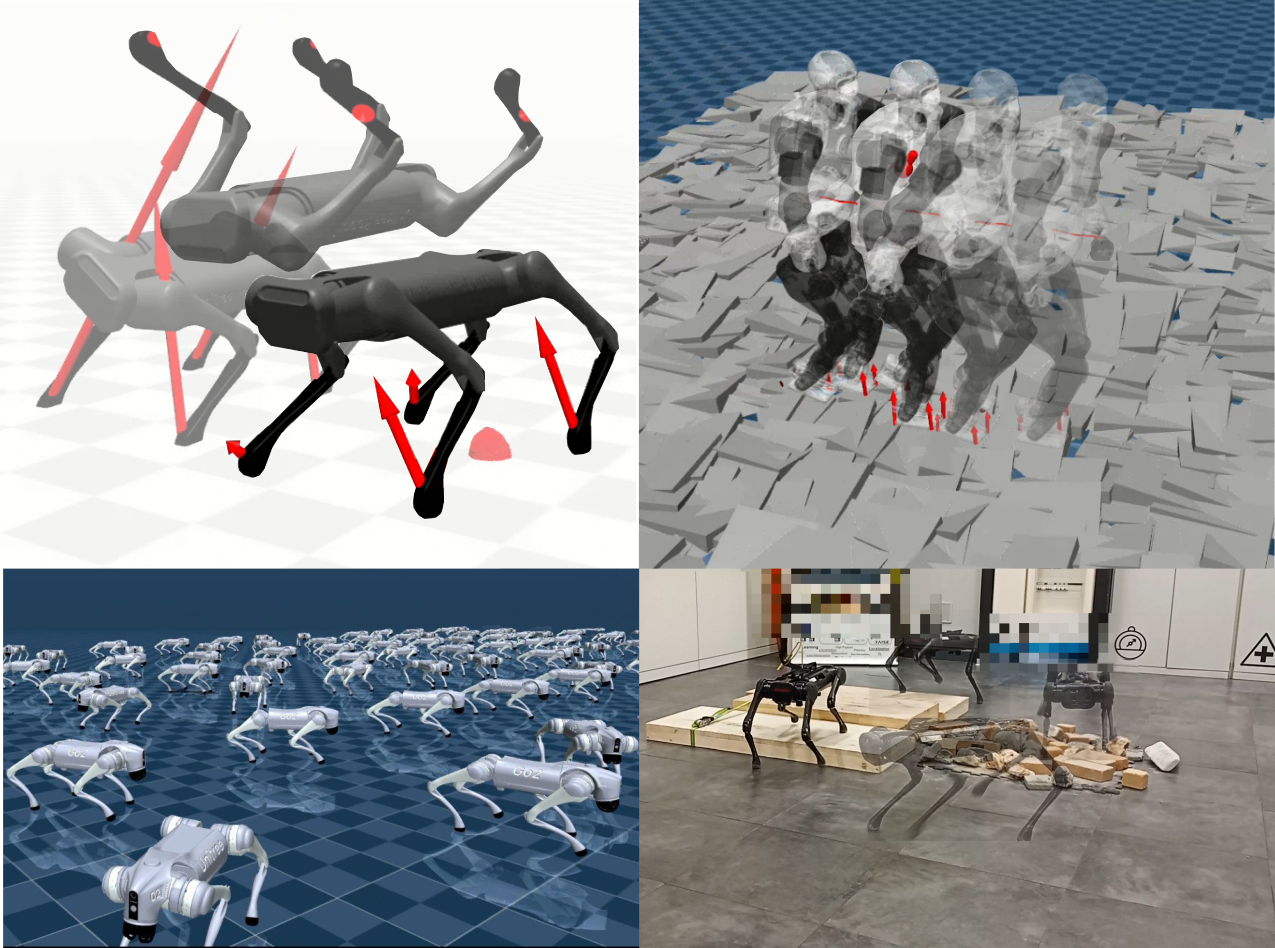}
         \caption{Top left: the quadruped robot Aliengo performing a barrel roll.
         Top right: the humanoid robot Talos walking blindly on uneven terrain. 
         Bottom left: an example of the massive parallelization achievable with our JAX formulation.
         Bottom right: Aliengo robot walking blindly on uneven terrain in a lab experiment.}
\label{fig:examples}
\vspace{-20pt}
\end{figure}
Another common approach to the solution of the \gls{ocp} is \gls{sqp}. Compared to \gls{ddp}-like algorithms, \gls{sqp} has been more extensively developed for general-purpose solvers due to its flexibility in handling a broader range of constraints and objectives, as well as its robustness in dealing with infeasible iterations. Solvers like OCS2 \cite{ocs2} and acados \cite{acados} also showed relevant results on real hardware as demonstrated in \cite{tamols} and \cite{mpc_acados}, thanks to the tailoring of the solver to the structure of \gls{ocp}. Recent work like \cite{sqp_lqr} highlights the connection between \gls{sqp} and \gls{ddp}, describing how the sparsity structure of the \gls{ocp} can be exploited not only in \gls{ddp}-like solvers but also in other solver types, such as \gls{sqp} or interior point-based solvers. 

While the above methods have been shown to exploit the structure of the \gls{ocp}, they have struggled to exploit new hardware accelerators like GPUs. Compared to CPUs, graphics cards can massively parallelize computation and are specifically designed for high-throughput linear algebra operations. Jeon et al. \cite{cusadi} tried to bridge the gap by developing Cusadi, a tool to convert expressions written in the well-known framework CasADI \cite{casadi} to GPU. Cusadi only translates closed-form expressions into GPU-compatible code, lacking the branching and looping capabilities necessary for most solvers. Bishop et al., \cite{reluQP} introduced a GPU accelerated \gls{qp} solver, Relu-QP, leveraging the \gls{admm} algorithm, achieving performance superior to state-of-the-art CPU-based solutions. However, its applicability to \gls{nlp} is constrained by the need to pre-compute part of the algorithm offline. Lee et al. \cite{gpu_lqr_limited_parallelization} implements an \gls{ilqr} controller on GPU but only partially leverages its capabilities, limiting parallelization to the line search and gradient computation. In contrast, \cite{gpu_lqr_batched} focuses only on batching, solving multiple \gls{ocp}s in parallel without exploiting state or temporal parallelism. \cite{gpu_ddp} extends the approach of \cite{gpu_lqr_limited_parallelization} by incorporating the temporal parallelization strategy introduced in \cite{Mpc}. However, \cite{Mpc} requires introducing approximations in the backward pass and performing a consensus sweep to maintain consistency. 
\revs{Frasch et al. \cite{moritz_parallel}, proposed a method to iteratively solve linear quadratic control exploiting parallel computation, while Wright et al. \cite{partitioned_dynamics} showed how to partition dynamic programming for parallel computation. However, none of the mentioned methods achieved at the same time logarithmic time complexity and an exact formulation. On the other hand,} Särkkä et al. \cite{TemporalParallelizationLQR} showed how parallel associative scan operations can be utilized to improve the computational complexity of LQR. When parallelized on a GPU, associative scans can compute the optimal control policy in $\mathcal{O}(\log^2(n)\log{N} + \log^2(m))$ instead of the $\mathcal{O}(N (n+m)^3)$ of the classical Riccati Recursion, where $n$ and $m$ are the dimensions of the system state and control input, and $N$ is the prediction horizon length. 


In this work, we designed \revf{an} \gls{sqp} method that uses an associative scan-based LQR solver to solve the primal-dual Karush-Kuhn-Tucker (KKT) system efficiently. The solver avoids the offline precalculation necessary for Relu-QP \cite{reluQP}. Our method benefits from a multiple shooting implementation, instead of the single shooting approach presented in \cite{gpu_ddp}. Our algorithm fully exploits the parallelization capabilities of GPUs in the temporal, state, and control dimensions, while maintaining, in contrast to \cite{gpu_ddp}, an exact backward pass.  We tailored the formulation for the deployment as a receding horizon controller for legged robot locomotion and analyzed its impact on learning and control.

\textit{Control:}~~At first, the reduction in the computational complexity of the solver may appear to have limited significance in a practical application. Some of the already described methods, like  OCS2 \cite{ocs2} and crocoddyl \cite{crocoddyl}, have been successfully used for whole-body \gls{mpc} even for experiments, as shown in \cite{inverse_dynamics_mpc} and \cite{Perceptive_based_MPC}. 
To ensure that such an \gls{mpc} formulation can run online on real hardware, the prediction horizon length and the number of robots that can be controlled simultaneously are constrained by the time complexity of the underlying algorithm. In particular, for solvers like OCS2, acados, and crocoddyl, the complexity scales linearly with the horizon length and cubically with the state dimension. In contrast, our approach scales with the square-log of the horizon length and logarithmically with the state and control dimensions. This improved scaling enables the use of more nodes at the same update rate, allowing for finer integration steps for a better approximation of the robot dynamic or, more critically, the inclusion of a \gls{mhe}, such as those presented by \cite{Invariant_moother} or \cite{inertial_estimationlocalization}, within the same optimization loop, opening the pace for "end-to-end" \gls{mpc}s capable of adapting and reacting to system changes directly using the sensor data. 
The reduction in complexity with respect to the state dimension enables better scalability across different robot morphologies. While other methods, such as the one presented by \cite{AcceleratingMPC}, use an approximate approach in the form of consensus \gls{admm} to parallelize over the system state, our implementation leverages the inherent parallelism of matrix operations provided by the GPU.
State space scaling becomes crucial when a centralized controller for collaborative tasks is considered. As demonstrated by \cite{CentralizedMPC} and \cite{LayeredMPC}, centralized controllers can handle complex tasks involving multiple collaborating agents. However, while these approaches achieve notable performances, the complexity of the task and the number of agents involved are constrained by their capability to solve the increasingly large \gls{ocp}s at a reasonable control frequency. Our framework overcomes such barriers, as shown in Sec. \ref{sub_seq::state_parallelization}. 

\textit{Learning:}~~In recent years, \gls{rl} has demonstrated remarkable performance, enhancing the robustness and capabilities of legged robots. \gls{rl} controllers have surpassed model-based controllers in robustness against model mismatches and sensor noise. Cheng et al. \cite{parkour} also demonstrated agile movements involving jumps on unstructured terrains with a real robot. However, \gls{rl} policies have shown limitations in scenarios that involve crossing gaps and stepping stones, since the learning process efficiency is significantly affected due to the sparse reward signals typical of such tasks. Giftthaler et al. \cite{DTC} proposed a solution to bridge the gap between learning-based and model-based controllers by incorporating \gls{mpc} as a bias for the \gls{rl}-policy. In contrast, works such as \cite{safesteps} and \cite{learning2closethegap} integrated learned behaviors directly into the optimization process.
Despite the noticeable results, all of the aforementioned methods that mix model-based controllers and \gls{rl} suffered from a slower training process due to the dependency on CPU-based solvers. In contrast, our implementation is entirely developed in JAX \cite{jaxgithub}, and it can be coupled with GPU-based simulators such as IsaacLab \cite{orbit} or Mujoco XLA \cite{mujoco} to achieve a significant speed-up in simulation throughput.
\subsection{Contribution}
To summarize, the major contributions of this work are:
\begin{itemize}
\item a novel GPU-accelerated MPC for legged robots that achieves logarithmic scaling in computation complexity on the horizon length, squared-logarithmic scaling in the state and control dimensions, and can easily be parallelized for use with data-driven approaches;
\item a detailed analysis of the performance benefits and limitations of the proposed algorithm with respect to state-of-the-art solvers; 
\item an open-source code repository for the rapid prototyping of MPC for legged robots and loco-manipulation, written in JAX; the code provides tools for controlling legged robots and also supports large-scale parallelization.
\end{itemize}

\subsection{Outline}
This paper is organized as follows. Sec. \ref{method} describes the details of the proposed solver. Sec. \ref{subsec::MPC} presents the dynamical models used in our formulation. Sec. \ref{sec::result} shows the benefits of our approach on comparative simulations against state-of-the-art solutions. Finally, Sec. \ref{sec::conclusion} draws the final conclusions.

\section{Primal-Dual LQR}
\label{method}
\subsection{Optimal Control Problem}
We start describing the \gls{ocp} at the base of the \gls{mpc} formulation we are presenting. In a \gls{ocp}, we transcribe not only the task we want to be performed but also the physical limits the system needs to respect. In this paper, we consider \gls{ocp}s of the following form:
\begin{subequations}
    \label{eq::ocp}
    \begin{align}
        \min\limits_{\boldsymbol{x},\boldsymbol{u}} \quad &\boldsymbol{l}_{N+1}(\boldsymbol{x}_{N+1})+\sum_{i = 0}^{N}{\boldsymbol{l}_i(\boldsymbol{x}_i ,\boldsymbol{u}_i)}\\
         s.t. \quad &\boldsymbol{x}_{i+1} = \boldsymbol{h}_i(\boldsymbol{x}_i,\boldsymbol{u}_i) \label{sub_eq::ocp_dynamcis} \\
         &\boldsymbol{u}_i \; \in \; \text{U}_i \label{sub_eq::ocp_uk}\\
         &\boldsymbol{x}_i \; \in \; \text{X}_i\label{sub_eq::ocp_xk}\\
        &\boldsymbol{x}_{0} = \boldsymbol{\hat{x}}_{0}
        \label{sub_eq::ocp_x0}
    \end{align}
\end{subequations}
where \revf{$\boldsymbol{x}_i $} is the system state and \revf{$\boldsymbol{u}_i$} is the control input. \revf{$\boldsymbol{l}(\boldsymbol{x_i},\boldsymbol{u_i})$} \revf{is the stage cost made of a quadratic tracking and regularization terms, while $\boldsymbol{l}(\boldsymbol{x_N})$ is the terminal cost}\revf{, and $h(\boldsymbol{x}_i,\boldsymbol{u}_i)$ is the system dynamics}.
Finally, $\text{X}_i$ and $\text{U}_i$ are, respectively, the set of feasible system states and control inputs.
\label{sec::primalduallqr}
The remainder of this section will describe the methods we use to solve the optimization problem posed by the \gls{ocp} formulation presented in \eqref{eq::ocp}.
By leveraging both the knowledge of the problem structure and exploiting modern hardware accelerators, we are able to obtain logarithmic time complexity in horizon length $N$ as well as square-log for the state $x$ and input $u$ dimensions. For simplicity, in this section, we avoid treating the inequality constraint in \eqref{sub_eq::ocp_xk} and \eqref{sub_eq::ocp_uk}. These details will be treated separately in Sec. \ref{sec:practical_implementation}. We now focus on the solution of the \revf{equality-constrained} optimization problem, which only presents the initial condition \eqref{sub_eq::ocp_x0} and dynamics  \eqref{sub_eq::ocp_dynamcis} as constraints. In particular, we are going to derive a multiple-shooting approach, which, in contrast to the single-shooting one, keeps both the state and control as optimization variables. We first derive an efficient \gls{sqp} algorithm that exploits the Riccati recursion to solve the equality constrained \gls{qp} Sec. \ref{sec::lqr}, and then we focus on the temporal parallelization of the solver through the use of parallel associative scan, Sec. \ref{sec::scan}. 
\subsection{Sequential Quadratic Programming}
We start by defining the Lagrangian of problem in \eqref{eq::ocp} as
\begin{equation}
\begin{aligned}
     \mathcal{L}(\boldsymbol{x},\boldsymbol{u},\boldsymbol{\lambda}) = 
     &\boldsymbol{l}_{N+1}(\boldsymbol{x}_{N+1}) + \boldsymbol{\lambda}_0^T(\boldsymbol{\hat{x}}_{0} - \boldsymbol{x}_0)
     +\\
     &\sum_{i = 0}^{N}{\boldsymbol{l}_i(\boldsymbol{x}_i ,\boldsymbol{u}_i)}+
     \boldsymbol{\lambda}_{i+1}^T(\boldsymbol{h}_i(\boldsymbol{x}_i,\boldsymbol{u}_i) - \boldsymbol{x}_{i+1}),
\end{aligned}
\end{equation}
where $\boldsymbol{\lambda}_i$ for $i = 1 \,\ldots \, N+1$ are the Lagrangian multipliers associated with the dynamics constraints (also known as costates). In an \gls{sqp} method\revf{,} we iteratively solve the \gls{qp} derived from the quadratic approximation of the Lagrangian and the linearization of the constraints at the current guess, $\boldsymbol{x}^k$, $\boldsymbol{u}^k$, where $k$ represents the last iterate. The \gls{qp} is then written as
\begin{equation}
\label{eq:QP}
\begin{aligned}
    \min_{\boldsymbol{\delta x}, \boldsymbol{\delta u}} & \sum_{i=1}^{N} 
    \begin{bmatrix}
        \boldsymbol{q}_i \\ \boldsymbol{r}_i
    \end{bmatrix}^\top 
    \begin{bmatrix}
        \boldsymbol{\delta x_i} \\ \boldsymbol{\delta u_i}
    \end{bmatrix} 
    + \frac{1}{2}
    \begin{bmatrix}
        \boldsymbol{\delta x_i} \\ \boldsymbol{\delta u_i}
    \end{bmatrix}^\top
    \begin{bmatrix}
        \boldsymbol{Q}_i & \boldsymbol{S}_i^\top \\
        \boldsymbol{S}_i & \boldsymbol{R}_i
    \end{bmatrix}
    \begin{bmatrix}
        \boldsymbol{\delta x}_i \\ \boldsymbol{\delta u}_i
    \end{bmatrix} \nonumber \\
    & + \boldsymbol{p}_{N+1}^\top \boldsymbol{\delta x}_{N+1} 
    + \frac{1}{2} \boldsymbol{\delta x}_{N+1}^\top \boldsymbol{P}_{N+1} \boldsymbol{\delta x}_{N+1}\\
    &\boldsymbol{\delta x}_0 = \boldsymbol{\hat{x}}_{0} - \boldsymbol{x}_0^k \\
    &\boldsymbol{\delta x}_{i+1} = \boldsymbol{A}_i \boldsymbol{\delta x}_i + \boldsymbol{B}_i \boldsymbol{\delta u}_i + \boldsymbol{b}_i ~~\text{for} \quad i = \revf{0}, \dots, N,
\end{aligned}
\end{equation}
where the search directions $\boldsymbol{\delta x} $ and $\boldsymbol{\delta u}$ are defined as


\begin{equation}
\begin{aligned}
    \delta \boldsymbol{x}_i &= \boldsymbol{x}^{k+1}_i - \boldsymbol{x}^k_i ~~~~~\text{for}\quad i = 0,\, \ldots, \, N + 1 \\
    \boldsymbol{\delta u}_i &= \boldsymbol{u}^{k+1}_i - \boldsymbol{u}^k_i ~~~~~\text{for}\quad i = 0,\, \ldots, \, N.
\end{aligned}
\end{equation}
The linearized dynamics is defined as
\[
\boldsymbol{A}_i = \frac{\partial \boldsymbol{h}_i}{\partial \boldsymbol{x}_i}, \,
\boldsymbol{B}_i = \frac{\partial \boldsymbol{h}_i}{\partial \boldsymbol{u}_i}, \,
\boldsymbol{b}_i = \boldsymbol{h}_i - \boldsymbol{x}^k_{i+1}.
\] 
The linear terms in the objective function are defined as
\[
\boldsymbol{q}_i = \nabla_{\boldsymbol{x_i}}\mathcal{L}, \quad
\boldsymbol{r}_i = \nabla_{\boldsymbol{u_i}} \mathcal{L}, \quad
\boldsymbol{p}_{N+1} = \nabla_{\boldsymbol{x}_{N+1}} \mathcal{L}.
\]
Finally, the Quadratic terms are defined as
\[
\boldsymbol{P}_{N+1} = \nabla^2\mathcal{L}_{N+1}, \quad
\begin{bmatrix}
    \boldsymbol{Q}_i & \boldsymbol{S}_i^\top \\
    \boldsymbol{S}_i & \boldsymbol{R}_i
\end{bmatrix}
= \nabla^2\mathcal{L}_i,
\]
where we removed the dependency of $\boldsymbol{x}^k_i$ and $\boldsymbol{u}^k_i$ from the dynamics $\boldsymbol{h}_i$ and the cost function $\boldsymbol{l}_i$. 
\subsection{Primal \& dual problem solution}
\label{sec::lqr}
As shown by \cite{fastGeneration}, such QPs can be efficiently solved by exploiting the problem structure. By applying the Riccati Recursion, starting from the last (i.e. with index $N$) node in the horizon and going backward in time, we obtain: 
\begin{equation}
\label{eq::lqr}
\begin{aligned}
    \boldsymbol{G}_i &= \boldsymbol{R}_i + \boldsymbol{B}_i^\top \boldsymbol{P}_{i+1} \boldsymbol{B}_i, \\
    \boldsymbol{H}_i &= \boldsymbol{S}_i + \boldsymbol{B}_i^\top \boldsymbol{P}_{i+1} \boldsymbol{A}_i, \\
    \boldsymbol{h}_i &= \boldsymbol{B}_i^\top (\boldsymbol{p}_{i+1} + \boldsymbol{P}_{i+1} \boldsymbol{b}_i) + \boldsymbol{r}_i, \\
    \boldsymbol{K}_i &= -\boldsymbol{G}_i^{-1} \boldsymbol{H}_i, \\
    \boldsymbol{k}_i &= -\boldsymbol{G}_i^{-1} \boldsymbol{h}_i, \\
    \boldsymbol{P}_i &= \boldsymbol{Q}_i + \boldsymbol{A}_i^\top \boldsymbol{P}_{i+1} \boldsymbol{A}_i + \boldsymbol{K}_i^\top\boldsymbol{H}_i, \\
    \boldsymbol{p}_i &= \boldsymbol{q}_i + \boldsymbol{A}_i^\top (\boldsymbol{p}_{i+1} + \boldsymbol{P}_{i+1} \boldsymbol{b}_i) + \boldsymbol{K}_i^\top \boldsymbol{h}_i.
\end{aligned}
\end{equation}
where $\boldsymbol{K}_i$ and $\boldsymbol{k}_i$ are, respectively, the feedback and feed-forward term of the optimal control policy. We can now retrieve the control and state vectors by recursively applying the forward pass starting from the initial condition as follows:
\begin{equation}
    \boldsymbol{\delta u}_{i} = \boldsymbol{K}_i\boldsymbol{\delta x_i}+\boldsymbol{k}_i,  \quad
    \boldsymbol{\delta x}_{i+1} = \boldsymbol{A}_i\boldsymbol{\delta x}_i+\boldsymbol{B}_i\boldsymbol{\delta u}_i+\boldsymbol{b}_i.
\end{equation}
Once we retrieve the values of $\boldsymbol{P}_i$ and $\boldsymbol{p}_i$ for $i = 0, \dots,N+1$, we can compute the update of the Lagrangian multiplier $\boldsymbol{\delta \lambda}$  in parallel. This is achieved by applying, at each node, the following update:
\begin{equation}
\boldsymbol{\delta \lambda}_i = \boldsymbol{P}_i\boldsymbol{\delta x}_i + \boldsymbol{p}_i.
\end{equation}
\subsection{Parallel Associative Scan}
\label{sec::scan}
In this section, we will briefly recap how parallel associative scans can be used to solve the primal problem as shown for the first time by \cite{TemporalParallelizationLQR}. Given a sequence of elements $a_1, \dots, a_N$ and an associative operator $\otimes$ defined on them, a parallel associative scan can be used to evaluate the values $s_1, \dots, s_N$ as: \looseness-1
\begin{equation}
    (s_1,\;s_2,\;\dots,\;s_N) = (a_1,\;a_1\otimes a_2,\; \dots,\;a_1 \otimes a_2 \otimes \dots \otimes a_N)
    \label{scan}
\end{equation}
in $\mathcal{O}(\log{N})$ as shown by \cite{scan}. \revf{This is possible since $\otimes$ is an associative operation, allowing the computation to be reorganized into smaller interdependent subproblems.} To make use of this property, we first define the \gls{cvf} $V_{i \rightarrow{} j}(\boldsymbol{x}_i,\boldsymbol{x}_j)$ as the minimal cost related to the optimal trajectory that goes from state $\boldsymbol{x}_i$ at node $i$ in the horizon to state $\boldsymbol{x}_j$ at node $j$\revf{, with $i < j$.
Given the connection between the value function of \eqref{eq:QP} and its \gls{cvf}, $V_i(\boldsymbol{x}_i) = \max_{\boldsymbol{x}_{N+1}} V_{i \rightarrow N+1}(\boldsymbol{x}_i, \boldsymbol{x}_{N+1})$, we will use the scan to evaluate $V_i(\boldsymbol{x}_i)$ for $i = 0, \dots N+1$.
\revf{We begin by initializing the elements $a_i$ with the \gls{cvf} $V_{i\rightarrow{i+1}}$ defined between two adjacent states $\boldsymbol{x}_i$ and $\boldsymbol{x}_{i+1}$. Next, we define the operator $\otimes$ for dynamic programming as:
\begin{equation}
    V_{i \rightarrow{k}} \otimes V_{k \rightarrow{j}} = \min_{\boldsymbol{x}_k} \{ V_{i \rightarrow{k}} + V_{k \rightarrow{j}} \}
    \label{eq:otimesOperator}
\end{equation}
where  $\boldsymbol{x}_k$ is a third intermediate state between $i$ and $j$. Note that this operator is associative since the $\min$ operator is associative. We can now perform an associative scan on the \gls{cvf} to obtain $V_{i\rightarrow N+1}$ in a single pass, thus $V_i(\boldsymbol{x}_i)$ for all $i$ in the horizon. To derive the combination rule for \eqref{eq:otimesOperator}}, we start by noting that the \gls{cvf} of \eqref{eq:QP} is also a quadratic program  with affine equality constraint \eqref{sub_eq::ocp_dynamcis}, which can be represented in its dual form \cite{convex_optimization}} by introducing the Lagrangian multiplier $\boldsymbol{\eta}$:
\begin{equation}
\label{eq::conditional_value_function}
\begin{aligned}
     V_{i \rightarrow{} j}(\boldsymbol{x}_i,\boldsymbol{x}_j) = \max\limits_{\boldsymbol{\eta}} 
    &\frac{1}{2} \boldsymbol{x}_i^\top \boldsymbol{\tilde{P}}_{i , j} \boldsymbol{x}_i +
    \boldsymbol{\tilde{p}}_{i , j}^\top \boldsymbol{x}_i
    - \frac{1}{2} \boldsymbol{{\eta}}^\top \boldsymbol{\tilde{C}}_{i,j} \boldsymbol{{\eta}}\\ 
    &- \boldsymbol{{\eta}}^\top 
    \left( \boldsymbol{x}_j - \boldsymbol{\tilde{A}}_{i,j} \boldsymbol{x}_i - \boldsymbol{\tilde{b}}_{i , j} \right)
\end{aligned}
\end{equation}
where the tilde $\tilde{\cdot}$ is used to distinguish the values related to the conditional value function from the one in \eqref{eq::lqr}, while the subscripts are used to specify the initial and final state considered. It can be proved that, given $V_{i \rightarrow{k}}$ and $V_{k \rightarrow{j}}$ in the form of \eqref{eq::conditional_value_function}, $V_{i\rightarrow{j}}$ will still have the form of \eqref{eq::conditional_value_function} and is characterized by the combination rule
\begin{equation}
\label{eq::combination_rule}
\begin{aligned}
\boldsymbol{\tilde{P}}_{i,j} &= \boldsymbol{\tilde{A}}_{i , k}^\top \left(\boldsymbol{I} + \boldsymbol{\tilde{P}}_{k , j}\boldsymbol{\tilde{C}}_{i , k}\right)^{-1} \boldsymbol{\tilde{P}}_{k , j}\boldsymbol{\tilde{A}}_{i , k} + \boldsymbol{\tilde{P}}_{i , k}, \\
    \boldsymbol{\tilde{p}}_{i,j} &= \boldsymbol{\tilde{A}}_{k , j}^\top \left(\boldsymbol{I} + \boldsymbol{\tilde{P}}_{k , j}\boldsymbol{\tilde{C}}_{i , k}\right)^{-1} 
    \left(\boldsymbol{\tilde{p}}_{k , j} - \boldsymbol{\tilde{P}}_{k , j}\boldsymbol{b}_{i , j}\right) + \boldsymbol{\tilde{p}}_{i , k}, \\
    \boldsymbol{\tilde{A}}_{i, j} &= \boldsymbol{\tilde{A}}_{k , j} \left(\boldsymbol{I} + \boldsymbol{\tilde{C}}_{i , k}\boldsymbol{\tilde{P}}_{k , j}\right)^{-1} \boldsymbol{\tilde{A}}_{i , k}, \\
    \boldsymbol{\tilde{C}}_{i, j} &= \boldsymbol{\tilde{A}}_{k , j} \left(\boldsymbol{I} + \boldsymbol{\tilde{C}}_{i , k}\boldsymbol{\tilde{P}}_{k , j}\right)^{-1} \boldsymbol{\tilde{C}}_{i , k} \boldsymbol{\tilde{A}}_{k , j}^\top + \boldsymbol{\tilde{C}}_{k , j}, \\
    \boldsymbol{\tilde{b}}_{i, j} &= \boldsymbol{\tilde{A}}_{k , j} \left(\boldsymbol{I} + \boldsymbol{\tilde{C}}_{i , k}\boldsymbol{\tilde{P}}_{k , j}\right)^{-1} 
    \left(\boldsymbol{\tilde{b}}_{i , k} - \boldsymbol{\tilde{C}}_{i , k}\boldsymbol{p}_{k , j}\right) + \boldsymbol{\tilde{b}}_{j , k}.
\end{aligned}
\end{equation}
We now initialize the values $a_i$ of a reverse associative scan, using the conditional value function $V_{i\rightarrow{i+1}}$ characterized by
\begin{equation}
\label{eq:initialization}
   \begin{aligned}
    \boldsymbol{\tilde{A}}_{i,i+1} &= \boldsymbol{A}_i - \boldsymbol{B}_i \boldsymbol{R}_i^{-1} \boldsymbol{S}_i, &\boldsymbol{\tilde{P}}_{i,i+1} &= \boldsymbol{Q}_i - \boldsymbol{S}_i^\top \boldsymbol{R}_i^{-1} \boldsymbol{S}_i, \\
    \boldsymbol{\tilde{C}}_{i,i+1} &= \boldsymbol{B}_i \boldsymbol{R_i}^{-1} \boldsymbol{B}_i^\top, &\boldsymbol{\tilde{p}}_{i,i+1} &= \boldsymbol{q}_i - \boldsymbol{S}_i^\top \boldsymbol{R}_i^{-1} r_i, \\
    \boldsymbol{\tilde{b}}_{i,i+1} &= \boldsymbol{b}_i - \boldsymbol{B}_i \boldsymbol{R}_i^{-1} \boldsymbol{r}_i,
    \end{aligned}
\end{equation}
and, in the case of $V_{N , N+1}$, by
\begin{equation}
   \begin{aligned}
    \boldsymbol{\tilde{P}}_{N , N+1} &= \boldsymbol{Q}_{N+1},
    &\boldsymbol{\tilde{A}}_{N , N+1} &= 0 , &\boldsymbol{\tilde{b}}_{N , N+1} = 0, \\
    \boldsymbol{\tilde{p}}_{N , N+1} &= \boldsymbol{q}_{N+1},
    &\boldsymbol{\tilde{C}}_{N , N+1} &= 0 .
   \end{aligned}
\end{equation}
From the solution of the parallel associative scan using the combination rule in \eqref{eq::combination_rule}, we obtain $V_{i \rightarrow N+1}(\boldsymbol{x}_i, \boldsymbol{x}_{N+1})$ for $i = 0,\dots,N+1$.
Finally, recalling that $\max_{\boldsymbol{x}_{N+1}} V_{i\rightarrow{N+1}}(\boldsymbol{x}_i,\boldsymbol{x}_{N+1})$ is equal to the value function $V_i(x_i)$ of the problem in \eqref{eq:QP} and that the value function is also in the same form of \eqref{eq::conditional_value_function}, we get that $ \boldsymbol{P}_i =  \boldsymbol{\tilde{P}}_{i,N+1}$ and $ \boldsymbol{\tilde{p}}_{i} =  \boldsymbol{p}_{i,N+1}$. This means that we obtained all $\boldsymbol{P}_i$ and $\boldsymbol{p}_i$ for $i = 0 , \, \dots \, ,N+1$ that can then be used in \eqref{sec::lqr} to calculate $\boldsymbol{K}$ and $\boldsymbol{k}$ in on pass in parallel. 
Thanks to a GPU-accelerated scan implementation, we can obtain the control policy over the entire horizon with an overall computational complexity of  \revf{$\mathcal{O}(\log^2(n)\log{N} + \log^2(m))$. This follows from the fact that both computing and initializing the scan involve a matrix inversion, which can be performed in parallel with complexity $\mathcal{O}(\log^2(n))$  \cite{parallel_matrix_inv}.}
The parallel associative scan can also be used to retrieve the optimal trajectory in $\mathcal{O}(\log{n}\log{N} + \log{m})$. We start by plugging into the linearized dynamics the optimal control law we evaluated, obtaining
\begin{equation}
\label{eq::controlled_dynamics}
    \boldsymbol{\delta x}_{i+1} = \boldsymbol{\bar{A}}_i\boldsymbol{\delta x}_i + \boldsymbol{\bar{b}}_i,
\end{equation}
where $\boldsymbol{\bar{A}}_i = \boldsymbol{A}_i+\boldsymbol{B}_i\boldsymbol{K}_i $ and $\boldsymbol{\bar{b}}_i = \boldsymbol{B}_i\boldsymbol{k}_i+\boldsymbol{b}_i$.
We can define the conditional optimal trajectory as follows:
\begin{equation}
\label{eq::conditional_optimal_trajectory}
\boldsymbol{\bar{h}}_{i\rightarrow{j}}(\boldsymbol{\delta x}_i,\boldsymbol{\delta x}_j) =  \boldsymbol{\bar{A}}_{i,j}\boldsymbol{\delta x}_i + \boldsymbol{\bar{b}}_{i,j}.
\end{equation}
\cite{TemporalParallelizationLQR} showed that from the combination of two conditional optimal trajectory $\boldsymbol{h}_{i\rightarrow{k}}$ and $\boldsymbol{h}_{k\rightarrow{j}}$ we obtain $\boldsymbol{h}_{i\rightarrow{j}}$ that can be written in the same form as \eqref{eq::conditional_optimal_trajectory} with
\begin{equation}
    \boldsymbol{\bar{A}}_{i,j}= \boldsymbol{\bar{A}}_{i,k}\boldsymbol{\bar{A}}_{k,j} \quad
    \boldsymbol{\bar{b}}_{i,j}=\boldsymbol{\bar{A}}_{k,j}\boldsymbol{\bar{b}}_{i,k}+\boldsymbol{\bar{b}}_{k,j}.
\end{equation}
We can initialize the element of parallel scan $a_i = \boldsymbol{h}_{i\rightarrow{i+1}}(\boldsymbol{x}_i)$ as
\begin{equation}
    \boldsymbol{\bar{A}}_{i,i+1}= \boldsymbol{\bar{A}}_{i}, \quad \boldsymbol{\bar{b}}_{i,i+1}= \boldsymbol{\bar{b}}_{i},\\
\end{equation}
for $i = 1, \dots, N$, while for $i= 0$ we set $\boldsymbol{\bar{A}}_{0,1} = \boldsymbol{0}$ and $\boldsymbol{\bar{b}}_{0,1} = \boldsymbol{\bar{A}}_0\boldsymbol{\delta x}_0+\boldsymbol{\bar{b}}_0$. From the parallel scan, we obtained the updated optimal state as:
\begin{equation}
    \boldsymbol{\delta x}_i = a_0 \otimes a_1 \otimes \cdots \otimes a_{i-1}
\end{equation}
\subsection{ Parallel Line Search}
Once we update the optimal directions $\boldsymbol{\delta x}$, $\boldsymbol{\delta u}$, and $\boldsymbol{\delta \lambda}$, we can perform a backtracking line search to calculate the step length $\alpha$ to finally get the new guess as
\begin{equation}
    \label{eq::linear_rollout}
    \begin{aligned}
        &\boldsymbol{x}^{k+1} = \boldsymbol{x}^{k} + \alpha\boldsymbol{\delta x},  &\boldsymbol{\lambda}^{k+1} = \boldsymbol{\lambda}^{k} + \alpha\boldsymbol{\delta \lambda},\\
        &\boldsymbol{u}^{k+1} = \boldsymbol{u}^{k} + \alpha\boldsymbol{\delta u}. &
    \end{aligned}
\end{equation}
The \gls{sqp} linear rollout allows for the computation to be easily performed in parallel for the horizon. To evaluate $\alpha$, we implemented a filter line search similar to the one implemented in \cite{ipopt}. The filter line search ensures that the step taken in the iteration reduces the constraint violation or the cost. Such a method has already been successfully deployed for legged robot control in \cite{Perceptive_based_MPC}. We measure the satisfaction \revf{of the dynamics constraints} as
\begin{equation}
\label{constraint satisfaction}
\theta  = \sum_{i=0}^{N+1}\|\boldsymbol{x}_{i+1} - \boldsymbol{h}(\boldsymbol{x}_i,\boldsymbol{u}_i)\|_2 .
\end{equation}
 As for \cite{Perceptive_based_MPC}, when the constraint violation $\theta$ is higher than a threshold, we reject the step if it further increases $\theta$. If the violation is below the threshold and the current step is a descent direction, then we apply an Armijo condition \cite{armijo}. On the other hand, if the current iterate is not in a descent direction, we require that at least the cost or $\theta$ is decreased to accept the step.
  \revf{To leverage GPU parallelism, we evaluate the line-search in parallel (at a fixed grid of ten step sizes, i.e.  $\alpha \in \{ 2^0, ..., 2^{-9}\}$) and select the largest $\alpha$ that satisfies one of the line-search conditions.}
\subsection{Practical Implementation}
\label{sec:practical_implementation}
As it is commonly done in practice, we also perform a Gauss-Newton approximation, ignoring the second-order term of the dynamics in \eqref{eq:QP}. We can further consider that in our case, the cost for the \gls{mpc} can be expressed as a non-linear least squares problem in the form $\frac{1}{2}\sum_{i = 0}^N \| \epsilon(x,u)\|_W$, where $W$ is a weight matrix. The Hessian approximation reduces to:
\begin{equation}
\begin{aligned}
    \boldsymbol{Q}_{N+1} &= \nabla_x \boldsymbol{\epsilon}_{N+1}^\top W_{N+1}\nabla_x \boldsymbol{\epsilon}_{N+1}  &\boldsymbol{S}_i =  \nabla_x \boldsymbol{\epsilon}_i^\top W_i\nabla_u \boldsymbol{\epsilon}_i  \\
    \boldsymbol{Q}_i &=  \nabla_x \boldsymbol{\epsilon}_i^\top W_i\nabla_x \boldsymbol{\epsilon}_i 
    &\boldsymbol{R}_i = \nabla_u \boldsymbol{\epsilon}_i^\top W_i\nabla_u \boldsymbol{\epsilon}_i
\end{aligned}
\end{equation}
Such an approximation also guarantees that \ref{eq:QP} is convex. As described in Sec. \ref{sec::primalduallqr}, the inequality constraints in \eqref{sub_eq::ocp_uk} and \eqref{sub_eq::ocp_xk} were not explicitly included in the solver formulation. However, these constraints are critical for \gls{mpc} as they represent the physical limitations of the system, such as friction cones or joint limits. To account for these limitations in the optimization, we incorporated them as a relaxed barrier function: \looseness-1
\begin{equation}
\mathcal{B}(\revf{\xi},\mu,\delta)
\begin{cases}
  -\mu \ln(\revf{\xi}), & \text{if } \revf{\xi} \geq \delta, \\
  \frac{\mu}{2} \left( \left( \frac{\revf{\xi} - 2\delta}{\delta} \right)^2 - 1 \right) - \mu \ln(\delta), & \text{if } \revf{\xi} < \delta,
\end{cases}
\end{equation}
where $\revf{\xi}(\boldsymbol{x},\boldsymbol{u}) < 0$ is the considered constraint, and $\mu$ and $\delta$ are parameters tuned for each constraint. Following the example of \cite{Perceptive_based_MPC}, to maintain the convexity of \ref{eq:QP} the Hessian approximation of the constraints is constructed as \begin{equation}
\begin{aligned}
    \boldsymbol{Q}_{N+1} &= \revf{ \nabla_x \boldsymbol{\epsilon}_{N+1}^\top W_{N+1}\nabla_x \boldsymbol{\epsilon}_{N+1}  } + \nabla_x \boldsymbol{\revf{\xi}}_{N+1}^\top \nabla_{\revf{\xi}\revf{\xi}}\mathcal{B}_{N+1}\nabla_x \boldsymbol{\revf{\xi}}_{N+1}  \\
    \boldsymbol{Q}_i &=  \revf{\nabla_x \boldsymbol{\epsilon}_i^\top W_i\nabla_x \boldsymbol{\epsilon}_i} + \nabla_x \boldsymbol{\revf{\xi}}_i^\top  \nabla_{\revf{\xi}\revf{\xi}}\mathcal{B}_i\nabla_x \boldsymbol{\revf{\xi}}_i  \\
    \boldsymbol{S}_i &= \revf{\nabla_x \boldsymbol{\epsilon}_i^\top W_i\nabla_x \boldsymbol{\epsilon}_i} + \nabla_x \boldsymbol{\revf{\xi}}_i^\top  \nabla_{\revf{\xi}\revf{\xi}}\mathcal{B}_i\nabla_u \boldsymbol{\revf{\xi}}_i  \\
    \boldsymbol{R}_i &= \revf{\nabla_u \boldsymbol{\epsilon}_i^\top W_i\nabla_u \boldsymbol{\epsilon}_i} + \nabla_u \boldsymbol{\revf{\xi}}_i^\top  \nabla_{\revf{\xi}\revf{\xi}}\mathcal{B}_i\nabla_u \boldsymbol{\revf{\xi}}_i
\end{aligned}
\end{equation}

As a common practice, while using the \gls{mpc} in closed-loop with the robot, we perform one iteration only of the algorithm described in Sec. \ref{sec::primalduallqr} per control loop. To increase the robustness of the method, we warm-start each call of the solver with the previous prediction just shifted by one time-step as shown by \cite{rti}.

Finally, to guarantee the differentiability of the solver through JAX-based automatic differentiation, we avoided the use of $while$ loop in the code. Especially in line-search, we evaluate a fixed number of alpha values in parallel. 

\section{Model Predictive Control}
\label{subsec::MPC}
In this section, we describe briefly the dynamics model $\boldsymbol{h}(\boldsymbol{x}_i,\boldsymbol{u}_i)$ presented in \eqref{sub_eq::ocp_dynamcis},
introducing two different models used for the formulation of the \gls{ocp} presented in the result Sec. \ref{sec::result}. The efficacy of the control input obtained from a \gls{mpc} strategy is directly related to the rate at which such a control strategy can be re-planned and to the accuracy of its predictions, as shown by \cite{solution_accuracy}. Therefore, we consider two main formulations from the literature: one based on the \gls{srbd} model and one based on the \gls{wb} model. 
The first approximates the system as \revf{a single body} with constant inertia, discarding the limb joint positions and the inertial variation from different limb configurations. This model is particularly effective in quadruped robots, where the limb mass is, in most cases, designed to be less than 10\% of the total mass \cite{cheeta_design}. On the other hand, the latter completely captures the dynamics of the robot, enabling the maximum exploitation of the system's capabilities. This benefit comes at the cost of handling a larger state space and highly non-convex models, thus affecting the maximum frequency at which the optimization problem can be solved. Moreover, the \gls{srbd}-\gls{mpc} state is limited to the center of mass linear and angular positions and velocities. This means that, when representing the robot’s orientation using quaternions, the state lies in $\mathbb{R}^7$. For such model, the control input consists only of the \gls{grf}, and thus lies in $\mathbb{R}^{3n_c}$, where $n_c$ is the number of contact points, \revf{meaning that, for the evaluations with the quadruped, in Sec. \ref{sec::result} $n$ is in $\mathbb{R}^{13}$ and $m$ in $\mathbb{R}^{12}$.}

In contrast, the \gls{wb}-\gls{mpc} also includes joint positions and velocities in the state, and joint torques in the control input. This increases the state dimensionality to $\mathbb{R}^{13 + 2n_{\text{joint}}}$ and the control input dimensionality to $\mathbb{R}^{3n_c + n_{\text{joint}}}$, \revf{that translate for the quadruped model in Sec. \ref{sec::result} in $n$ in $\mathbb{R}^{37}$ and $m$ in $\mathbb{R}^{24}$.}

Beyond the difference in dimensionality, the \gls{wb} model also introduces greater complexity. While the \gls{srbd} model features bilinear terms only in the angular dynamics, the \gls{wb} model has highly nonlinear dependencies on the joint configuration. As a result, the overall \gls{ocp} becomes significantly harder to solve. \looseness-1

\section{Results} \label{sec::result}
\begin{figure}[t]
         \centering
         \includegraphics[width=0.41\textwidth]{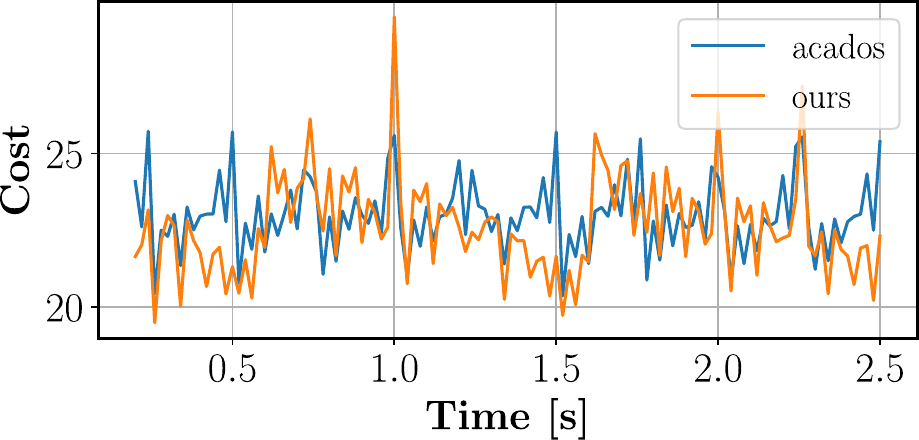}
         \caption{Comparison of the \gls{mpc} optimal cost, evaluated from the same initial condition with acados \cite{acados} (in blue) and with our implementation (in orange). Both solvers are evaluated at each call of the \gls{mpc} and the costs are recorded.}
    \label{fig:cost_comparison}
    \vspace{-5pt}
\end{figure}
\begin{figure}[t]
    \centering
    \includegraphics[width=0.41\textwidth]{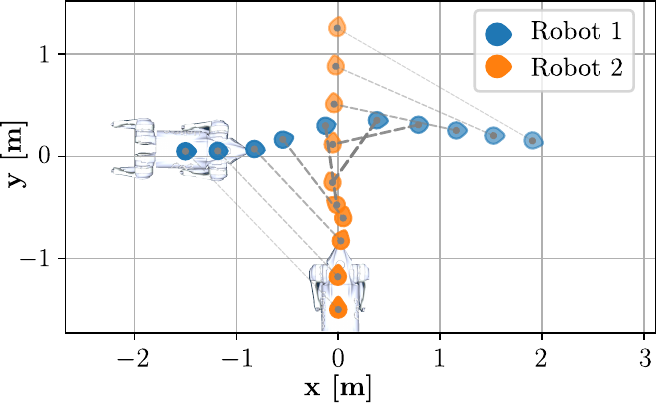}
    \caption{Path of the two robots tracking their desired trajectories that will lead to a collision. \revf{We visualize the distance between the two robots with dashed gray lines of different shades, the darker the smaller distance.}}
    \label{fig:centralized_trajectory}
    \vspace{-15pt}
\end{figure}
\begin{figure}[t]
         \centering
         \includegraphics[width=0.41\textwidth]{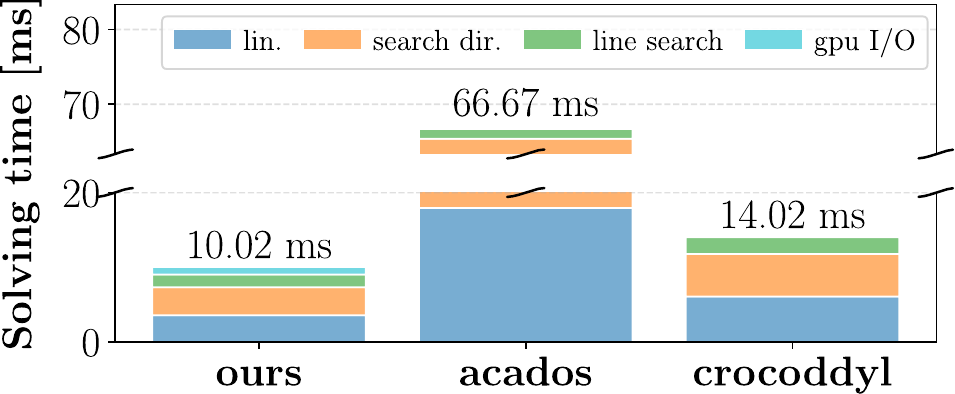}
         \caption{\revs{Breakdown of the solving times for our implementation, acados (SQP) and crocoddyl (FDDP).}}
    \label{fig:computation_breackDown}
    \vspace{-15pt}
\end{figure}
        
         
          
     
In this section, we present the results of various analyses that highlight the strength of the proposed approach. We evaluate the performance of our algorithm using the two models introduced in Sec. \ref{subsec::MPC} (the \gls{srbd} and \gls{wb} models). To benchmark our implementation against state-of-the-art solvers, we used acados \cite{acados} \revs{(\gls{sqp})} for comparisons on the \gls{srbd} model and crocoddyl \cite{crocoddyl} \revs{(DDP)} for the \gls{wb} model. We used acados for the \gls{srbd} comparison, as it could be easily adapted for the evaluations with multiple robots, shown in Fig. \ref{fig::centralized} and \ref{fig::multi_env}. On the other hand, we used crocoddyl for the horizon length comparison with the \gls{wb} model, as it showed significantly better performance than acados with the more complex model as shown in Fig. \ref{fig:computation_breackDown}. 

Our analysis includes investigating how the average solving time scales with the horizon length across all models and solvers. Additionally, we assess the solver's performance as the system state dimension increases, as well as the vectorization capabilities of our approach. \revf{All the solving time comparisons consider a single iteration for each algorithm.} The tests were conducted on a desktop computer equipped with an Intel Core i7-13700KF and an NVIDIA RTX 3080. We used Mujoco XLA, to evaluate the \gls{wb} dynamics. In all the presented benchmarks, if not stated differently, we used the 15 kg Unitree Go2 robot, a torque-controlled quadruped platform, \revf{with a horizon length N of 50 nodes}. As Fig. \ref{fig:examples} and the accompanying video show, our \gls{wb}-\gls{mpc} can also be implemented for real-time control of other robot morphologies, like humanoids.

\begin{figure}[!t]
\centering
    \includegraphics[width=0.41\textwidth]{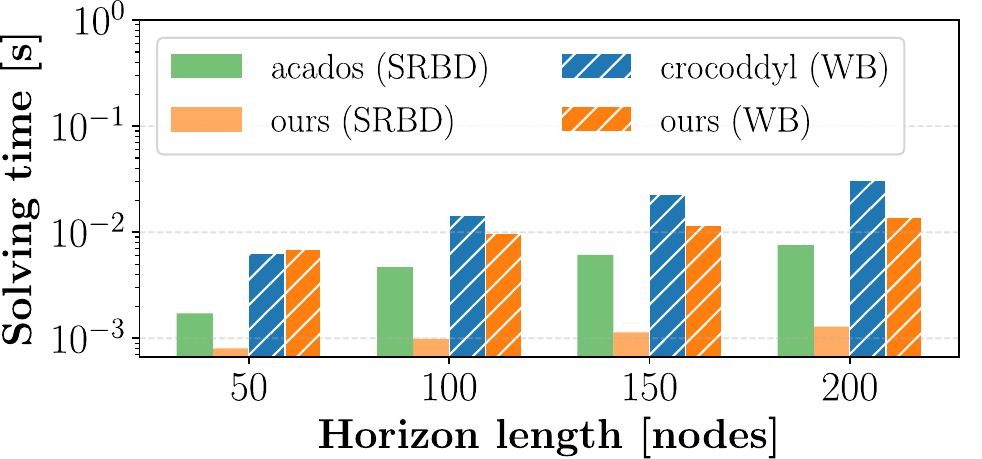}
    \caption{Average solving times when varying the horizon length (number of nodes) for the \gls{srbd} model (top) and \gls{wb} model (bottom). In blue and green, we plot the solution time of two different state-of-the-art solvers (respectively acados \cite{acados} and crocoddyl \cite{crocoddyl}), while in orange, we plot the one of our implementation.}
    \label{fig:horizon}
\vspace{-15pt}
\end{figure}

\subsection{Performance evaluation}
First, we evaluate the ability of our solver to generate complex dynamic maneuvers. In this scenario, we are generating a barrel roll motion with the Aliengo robot, as shown in Fig. \ref{fig:examples}. Thanks to its multiple shooting formulation, we can initialize the solver with an infeasible initial guess, and the solver converges in just $27$ iterations to a solution that completes the task.
We also assess the performance of the proposed solver in terms of solution optimality. In Fig. \ref{fig:cost_comparison}, we compared the values of the cost from the optimal solution given by our implementation against the one given by acados, which is chosen for this comparison because it allows using the same cost function as the one in our formulation. In this scenario, we are using the whole-body model and are controlling the robot in simulation at 50Hz. The robot is tasked to trot with a forward speed of $0.3$ m/s while being randomly pushed by an external disturbance of $50$N for $0.25$s. At every control loop, we record the value of the cost functional from both solvers, comparing solutions that start from the same initial condition. Figure \ref{fig:cost_comparison} reports the values recorded during a $2.5$s time span. Although the two solutions are comparably similar, our implementation results in a $20\%$ reduction in the average cost along the trajectory.

To finally validate the \gls{mpc}, we also performed experiments on the real hardware, as shown in Fig. \ref{fig:examples} and in the accompanying video. For the real experiments, the solver runs on an Nvidia RTX 4050 laptop GPU. Figure \ref{fig:examples} shows snapshots of the robot walking blindly up the steps and through uneven terrain using only proprioceptive data for state estimation \cite{muse}.
We also evaluated the proposed approach when used to control two different robots at the same time in a centralized control implementation in simulation. In this example, the two robots are controlled by the same \gls{mpc} that evaluates the control input for both systems at $50$Hz. The \gls{mpc} also includes a collision avoidance constraint, in the form of a quadratic penalty term. In the tested scenario, the two robots are tasked to track two perpendicular trajectories that cross each other and would lead to a collision. Figure \ref{fig:centralized_trajectory} shows the resulting path of the two robots\revf{, where the dashed grey lines represent the distance between the two robots}. As shown, both robots deviate from their desired paths to avoid collisions, demonstrating coordinated behavior. While being just a simple example, this result highlights the potential of our \gls{mpc} in solving collaborative tasks. Further implementation in more complex scenarios, such as collaborative carrying, is left for future work. \revf{Figure \ref{fig:computation_breackDown} presents the average solving time breakdown for the three solvers in the quadruped trotting scenario with a horizon of N=100.
It highlights the sequential backward pass in crocoddyl, the \gls{qp} solver call in acados, and the parallel scan in our method for evaluating the search direction.
A minor computational overhead of approximately $1ms$ arises from GPU communication, as only the initial state is sent and the commanded torques, joint positions, and velocities are retrieved.}

\subsection{Temporal Parallelization}
In this subsection, we show the benefit of our implementation against the state-of-the-art solvers acados \cite{acados} and crocoddyl \cite{crocoddyl}, in terms of average solving time for increasing horizon length. For this assessment, we initialized the robot in a \revf{feasible} random pose and joint configuration and tasked the controller to perform a forward trot at a speed of $0.5$m/s. Acados uses the same cost and dynamics formulation as in our controller, while for the comparison with crocoddyl we adapted our reference generator and cost to match the reference used in the example provided with crocoddyl's libraries. As reported in Fig. \ref{fig:horizon}, our implementation shows faster update rates than both solvers in almost all tested scenarios. For the assesment against acados, with the \gls{srbd} model, our implementation is always faster. For the case of the \gls{wb} formulation, crocoddyl is able to outperform our formulation with a prediction horizon smaller than $80$ nodes. This last outcome is primarily attributed to overhead involved in transferring data to and from the GPU, which, for smaller problems, can be detrimental. Noticeably, our formulation can achieve an update rate of 50Hz for a horizon length of 200 nodes. \looseness-1



\subsection{State Parallelization}
\label{sub_seq::state_parallelization}
\begin{figure}[!t]
\centering{
\includegraphics[width=0.41\textwidth]{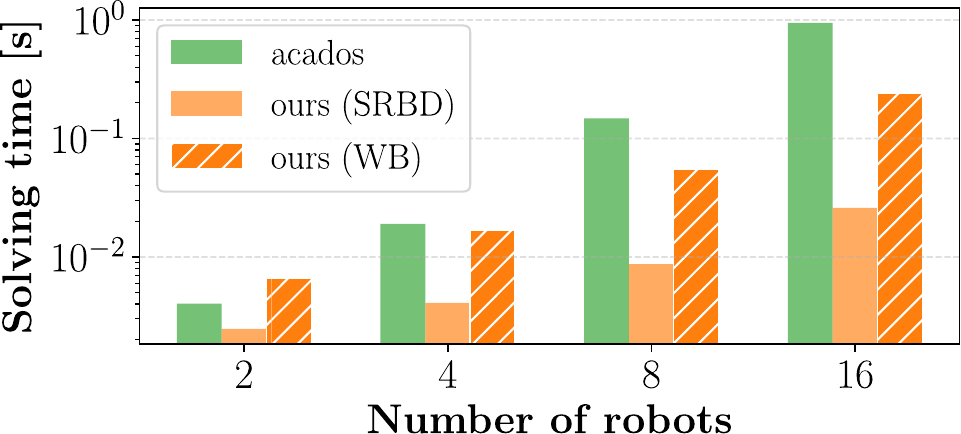}
}    
\caption{Our implementation can control multiple robots in the same optimization, a typical scenario found in centralized controllers. In green, we depict the solution time of acados, while in orange, ours, considering the \gls{srbd} model. Furthermore, in dashed orange, we report the result obtained by our implementation using the \gls{wb} model.}
\label{fig::centralized}
\vspace{-15pt}
\end{figure}
In this subsection, we discuss the benefits of our implementation in the context of centralized controllers. In our tests, we evaluated the average solving times for the \gls{mpc} while progressively increasing the number of robots included in the same optimization. This approach simulated a centralized controller managing $n$ different robots. For the \gls{srbd} model, we compared our implementation against acados. For the \gls{wb} model comparison, crocoddyl was excluded since its model interface does not allow for a customization that is needed for a fair evaluation. As Fig. \ref{fig::centralized} shows, our implementation outperformed acados when considering the \gls{srbd} model, achieving a computational time of 25 ms in the presence of 16 robots, enough for the real-time control of the systems. When using the \gls{wb} model, instead, computation becomes significantly more demanding for the implementation with acados and we could not record data for the comparison. As shown in Fig. \ref{fig::centralized}, our algorithm successfully controls up to four robots (\gls{wb} model case) at 20 Hz. To the best of the authors' knowledge, this level of performance has never been achieved before.\looseness-1
\begin{figure}[!t]
\centering{
\includegraphics[width=0.41\textwidth]{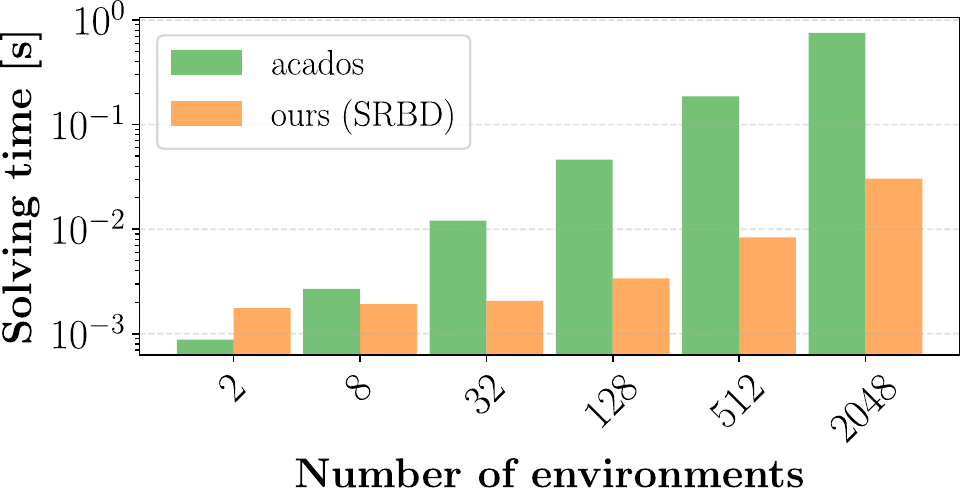}
}    
\caption{Comparison of the solving time achieved by our implementation against the batch-parallelized solver feature of acados, when varying the number of environments being simulated in parallel. In this example, each environment has a dedicated \gls{mpc}.}
\label{fig::multi_env}
\vspace{-15pt}
\end{figure}
\begin{table}[h!]
\centering
\rowcolors{2}{RowDark}{RowLight} 
\begin{tabular}{lccc} 
\toprule
 & $n_{\text{env}}$ & Control Frequency & Real-Time Factor \\
\midrule
\gls{srbd}-\gls{mpc} & 4096 & 50\,Hz & 370$\times$ \\
\gls{srbd}-\gls{mpc} & 4096 & 25\,Hz & 570$\times$ \\
\gls{wb}-\gls{mpc}   & 4096 & 50\,Hz & 35$\times$ \\
\gls{wb}-\gls{mpc}   & 4096 & 25\,Hz & 75$\times$ \\
\end{tabular}
\caption{Performance metric}
\label{tab:batch}
\vspace{-20pt}
\end{table}
\subsection{Model-Based Learning}

In this subsection, we highlight the vectorization capabilities of our controller. Thanks to its full implementation in JAX, we can easily vectorize the computation of the \gls{mpc}. Such ability is critical in the training scenario as shown in \cite{DTC} or at runtime as shown in \cite{mcts}. In Fig. \ref{fig::multi_env}, we compare our approach to the batched version of acados, which uses OpenMP to parallelize the computation of independent \gls{mpc} instances on the CPU. As the plot shows, acados is competitive with the proposed \gls{mpc} only for small batch sizes. This limitation comes from the reduced number of cores available on a CPU compared to the parallelization capabilities of modern GPUs.

Finally, Table \ref{tab:batch} summarizes the performance of our \gls{mpc} when batched across multiple environments with Mujoco XLA as the simulator. In this setup, we generate 4096 environments, each one running its own \gls{mpc}. For the \gls{srbd} model, we control the robot at an update rate of 50 Hz and we achieve a value of 570 seconds of simulated time per second. Meanwhile, for the \gls{wb} model, closing the loop at 25 Hz, we achieve a real-time factor of 75x.
\section{Conclusion} 
This work presents a novel GPU-accelerated \gls{mpc} framework implemented for legged robot locomotion. By leveraging the parallel processing capabilities of GPUs and implementing a Primal-Dual iLQR solver in JAX, we achieve logarithmic scaling in horizon length and square-log scaling with state and control dimensions. Our approach demonstrates significant improvements over state-of-the-art solvers, achieving higher computational efficiency and scalability. This allows for the optimization of centralized controllers for multiple robots and the integration of large-scale parallel environments, enhancing learning-based control frameworks.
Future works will include the exploitation of the presented \gls{mpc} as a bias in the learning process of a locomotion policy. Furthermore, we would like to include a more sophisticated methodology for dealing with inequality constraints.  
\label{sec::conclusion}

\vspace{-10pt}

\bibliographystyle{IEEEtran}
\bibliography{references}

\end{document}